\documentclass[conference]{IEEEtran}
\ifCLASSINFOpdf
\else
\fi
\usepackage{graphicx}
\graphicspath{{/}}

\begin{document}
%
\title{Scene Recognition by Combining Local and Global Image Descriptors}

\author{\IEEEauthorblockN{Jobin Wilson}
\IEEEauthorblockA{R\&D Department\\
Flytxt, India\\
jobin.wilson@flytxt.com}
\and
\IEEEauthorblockN{Muhammad Arif}
\IEEEauthorblockA{R\&D Department\\
Flytxt, India\\
muhammad.arif@flytxt.com}}


%


\maketitle

\begin{abstract}
Object recognition is an important problem in computer vision, having diverse applications.
In this work, we construct an end-to-end scene recognition pipeline consisting of feature extraction, encoding, 
pooling and classification. Our approach simultaneously utilize global feature descriptors as well as 
local feature descriptors from images, to form a hybrid feature descriptor corresponding to each image. 
We utilize DAISY features associated with key points within images as our local feature
descriptor and histogram of oriented gradients (HOG) corresponding to an entire image as a global descriptor. 
We make use of a bag-of-visual-words encoding and apply Mini-Batch K-Means algorithm to reduce the complexity 
of our feature encoding scheme. A 2-level pooling procedure is used to combine DAISY and HOG features corresponding to each image. 
Finally, we experiment with a multi-class SVM classifier with several kernels, in a cross-validation setting, and tabulate 
our results on the fifteen scene categories dataset. The average accuracy of our model was 76.4\% in the case of a 40\%\textendash60\% 
random split of images into training and testing datasets respectively. The primary objective of this work is to clearly outline 
the practical implementation of a basic screne-recognition pipeline having a reasonable accuracy, in python,
using open-source libraries. A full implementation of the proposed model is available in our github repository. 
\footnote{https://github.com/flytxtds/scene-recognition}

\end{abstract}


%
\IEEEpeerreviewmaketitle

\section{Introduction}
Object recognition is an interesting computer vision problem with diverse applications
in a variety of fields such as automated surveillance, robotics, human computer interaction, video indexing and 
vehicle navigation \cite{yilmaz2006object}. In this work, we build an end-to-end pipeline for recognizing natural scene categories 
in python and report our results on the fifteen scene categories dataset \cite{lazebnik2006beyond}. 
Our pipeline essentially consists of feature extraction, encoding, pooling and classification steps. 
However, we augment this standard pipeline by coming up with a hybrid feature descriptor that simultaneously 
make use of local feature descriptors as well as global feature descriptors at a different granularity, with the objective of
improving the classification accuracy. We also reduce the time-complexity of our encoding step by using Mini-Batch K-Means algorithm
instead of the standard K-means algorithm, for constructing the bag of visual words, since the number of feature descriptors 
involved would be quite large in practice. In our experiments with the fifteen scene categories dataset, the number of
feature descriptors  extracted from the training data set were around 1.9 million. We utilize a 2-level pooling scheme to combine 
DAISY and HOG features corresponding to each image.  We train a multi-class SVM classifier with multiple kernel choices in a 
cross-validation setting and tabulate our classification results.

We describe each of the steps in our pipeline in detail in Section~\ref{sec:methodology} and also present 
our design choices and their contributions in improving the overall performance of our scene recognition pipeline. 
In Section~\ref{sec:results}, we study the impact of the vocabulary size in our encoding step and the effects of various 
kernel choices as well as parameters, and present our results. We discuss the key insights in the light of our results in 
Section~\ref{sec:discussion}. Section~\ref{sec:conclusion} concludes our paper.

\section{Methodology}
\label{sec:methodology}
Our scene recognition pipeline involves feature extraction, encoding, pooling and classification steps. 
Our model has multiple tunable parameters such as size of the visual vocabulary ($K$), the kernel to be used by the SVM classifier 
(e.g. linear vs. Gaussian) as well as SVM hyper-parameters such as $C$ and $\gamma$. We make use of cross-validation to 
empirically choose the optimal hyper-parameters. We discover the optimal $K$ empirically for a linear SVM first. 
We use this $K$ and further perform a grid-search to discover the optimal $C$ and $\gamma$ corresponding to an SVM model having a Radial Basis Function(RBF) kernel. 
Details of each of our pipeline steps are presented below.

\subsection{Feature Extraction}
From each image, we first extract DAISY\cite{tola2008fast} features corresponding to the key-points detected from the image. 
We make use of the skimage library (scikit--image) \cite{van2014scikit} for feature extraction. Corresponding to each keypoint in 
the image, we extract a DAISY descriptor. We also extract a standard HOG descriptor corresponding to the whole image at a 
different granularity (by making use of the parameters pixels\_per\_cell, cells\_per\_block and orientations), effectively 
allowing us to choose features at different scales. We make use of the DAISY descriptors extracted from all the training images 
to form a bag-of-visual-words by clustering them using Mini-Batch K-Means. We empirically determine the best visual vocabulary 
size ($K$) through cross-validation. 

\subsection{Encoding}
We make use of the local DAISY features corresponding to each keypoints to encode each image as a histogram of visual words.
We use the standard ``bag-of-visual-words'' concept here. Concretely, we apply K-means algorithm to quantize 
DAISY features into `$K$' clusters to form ``visual words'' in a vocabulary. $K$ represents the vocabulary size. 
Corresponding to each image, we form a histogram with `$K$' as the dimensionality, by using this vocabulary.  This encoded 
representation of the image forms the ``DAISY histogram'' and constitutes a portion of our hybrid feature descriptor.

\subsection{Pooling}
We make use of a 2 level pooling scheme in our scene recognition pipeline. From the DAISY features, 
we construct a histogram by representing frequency of each visual word in each image. We perform this by selecting each 
key point in the image and looking up the cluster id corresponding to that DAISY descriptor and incrementing the count 
corresponding to that bin in our ``DAISY histogram''. This procedure is effectively a ``sum pooling''. We do an L2 
normalization of the resulting histogram to form a DAISY histogram feature which we call as ``DAISY histogram''. For the 2nd level 
pooling, we take the HOG global descriptor corresponding to each image and do an L2 normalization followed by a concatenation 
with the corresponding DAISY histogram feature, to form our hybrid feature descriptor.

\subsection{Classification}
We make use of the standard SVM classifier with various kernels, for classification. We utilize the sklearn library (sckit-learn) 
\cite{pedregosa2011scikit} for the SVM implementations. We perform crossvalidation by randomly splitting the dataset 
into a training and validation set. We build the ``visual vocabulary'' as well as training feature vectors from the training split. 
We report the overall accuracy, confusion-matrix as well as standard information-retrieval statistics such as precision, recall and 
f-measure corresponding to our experiments.

\section{Results}
\label{sec:results}
\begin{figure}
  \caption{Plot between $K$ and Accuracy(Linear SVM)}
  \label{fig:kmeans-accuracy}
  \centering
  \includegraphics[scale=0.5]{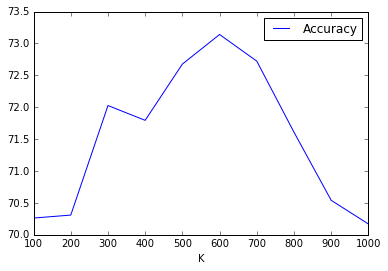}
\end{figure}

We conducted our experiments on a standard DELL Precision Tower 5810 workstation, running 64 bit ubuntu 14.04 LTS. 
Once the execution pipeline described in Section \ref{sec:methodology} was built, the optimal size of visual vocabulary ($K$) was 
emperically discovered, by running the pipeline 3 times corresponding to each $K$ and averaging their accuracy. 
A linear SVM model was used for classification. The optimal value of $K$ was found to be 600 and the average accuracy across 
3 runs was 73.14\%. The results are summarized in Table \ref{tab:kmeans-accuracy} and plotted in the Figure \ref{fig:kmeans-accuracy}
\begin{table}
  \centering
  \caption{Average accuracy of MiniBatch K-Means for different $K$}
  \label{tab:kmeans-accuracy}
  \begin{tabular}{|l|l|}
    \hline
    K & Accuracy(\%) \\ \hline
    100 &  70.2628 \\ \hline
      200 &  70.3092 \\ \hline
      300 &  72.0269 \\ \hline
      400 &  71.7948 \\ \hline
      500 &  72.6769 \\ \hline
      600 &  73.1411 \\ \hline
      700 &  72.7233 \\ \hline
      800 &  71.6091 \\ \hline
      900 &  70.5413 \\ \hline
     1000 &  70.1697 \\ \hline
  \end{tabular}
\end{table}

\begin{figure}
  \caption{RBF kernel optimization response surface}
  \label{fig:svm-tune-surface}
  \centering
  \includegraphics[scale=0.5]{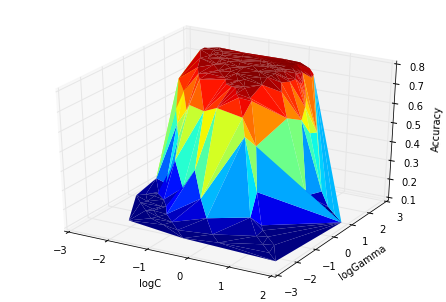}
\end{figure}

After obtaining the optimal $K$ using a linear SVM, we changed the kernel to RBF and performed a grid search for discovering the 
optimal $C$ and $\gamma$ by varying {\it log($C$)} within the interval [-3 3] and {\it log($\gamma$)} within the interval [-3 2] 
with the value of $K$ as 600. The optimal value of {\it log($C$)} was 1.679 and {\it log($\gamma$)} was -0.163 for 
an accuracy of 74.5\%. The optimization response surface is plotted in Figure \ref{fig:svm-tune-surface}.

\begin{figure}
  \caption{Confusion Matrix: SVM with hybrid feature and RBF Kernel (60\% Test data)}
  \label{fig:rbf-conf-mat}
  \centering
  \includegraphics[scale=0.5]{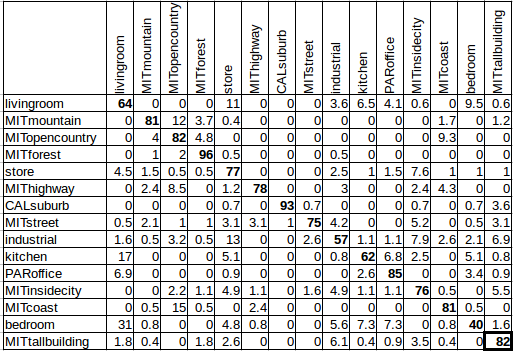}
\end{figure}

\begin{figure}
  \caption{Confusion Matrix: SVM with hybrid features and linear Kernel (60\% Test data)}
  \label{fig:lin-conf-mat}
  \centering
  \includegraphics[scale=0.5]{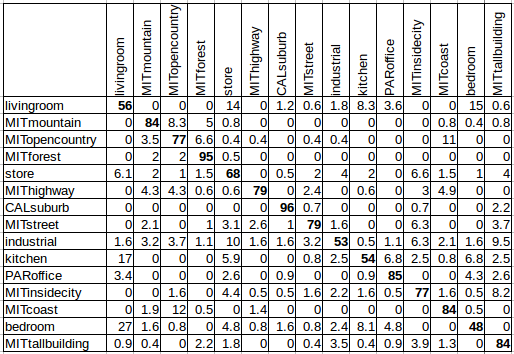}
\end{figure}

We executed the pipeline with an RBF kernel for optimal $K$, $C$ and $\gamma$, and the final accuracy was found to be 76.4\%.  
The average precision is 77\%, average recall is 76\% and average f1-score is 76\%. The confusion matrix for the model
is depicted in figure \ref{fig:rbf-conf-mat}. The pipeline with a linear kernel and optimal $K$ reported an accuracy 
of 75.6\%, average precision of 76\%, average recall of 76\% and average f1-score of 75\%. This shows that kernelizing 
SVM marginally improves the accuracy. The confusion matrix for the classifier is shown in Figure \ref{fig:lin-conf-mat}.

\begin{figure}
  \caption{Confusion Matrix: SVM with HOG feature alone and RBF Kernel (60\% Test data)}
  \label{fig:hog-conf-mat}
  \centering
  \includegraphics[scale=0.5]{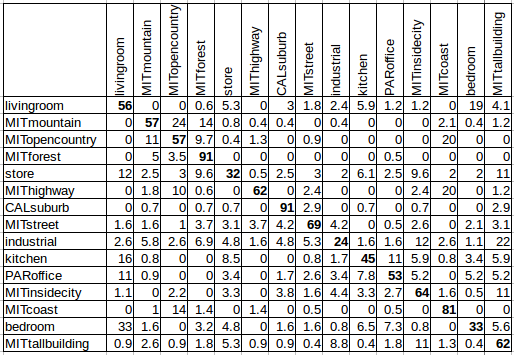}
\end{figure}

\begin{figure}
  \caption{Confusion Matrix: SVM with DAISY features alone and linear Kernel (60\% Test data)}
  \label{fig:daisy-conf-mat}
  \centering
  \includegraphics[scale=0.5]{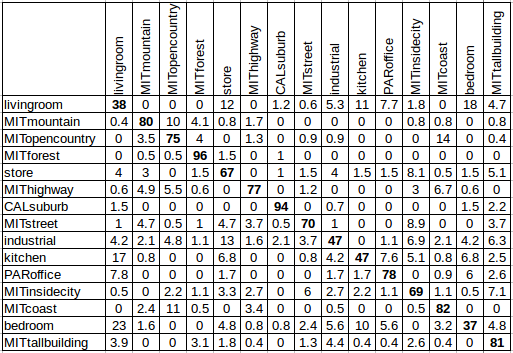}
\end{figure}

We observed that executing the pipeline(linear SVM) with HOG descriptors alone is providing an accuracy of 59.1\%, 
average precision of 59\%, average recall is 59\% and average f1-score is 58\%. With DAISY descriptors alone, 
the accuracy was 70.8\%, average precision was 70\%, average recall was 71\% and average f1-score was 70\%. 
The confusion matrix corresponding the model with HOG features alone and the model with DAISY features alone are shown 
in Figures \ref{fig:hog-conf-mat} and \ref{fig:daisy-conf-mat} respectively.
We observe that there is a significant increase in the model accuracy while employing a hybrid feature descriptor comprising
of HOG and DAISY descriptors, compared to models that utilize either HOG features or DAISY features only. 

\section{Discussion}
\label{sec:discussion}

We observe in section \ref{sec:results} that by utilizing a hybrid feature descriptor, our model accuracy is significantly improved. 
The relatively poor performance of models utilizing HOG descriptors alone can be attributed to its inability to deal with scale, 
orientation or translation. However, HOG descriptors can be combined with a dense feature descriptors such as DAISY which 
are capable of dealing with scale, orientation and translation, to produce superior results. 
The two-level pooling scheme used by our model improves the overall results. L2 normalization is performed on the DAISY 
and HOG descriptors to bring both these components to scale, so that their concatenation would result in a meaningful 
feature descriptor. We also used an approximation technique while building the bag-of-visual-words representation, to reduce the
algorithmic complexity involved in this process. In our experiments, the number of DAISY descriptors extracted from 
the training dataset was around {\it 1.9 million}. The scikit-learn K-Means implementation ran for several hours to produce a 
reasonable clustering. We made use of the Mini-Batch K-means proposed by David Sculley in his paper titled ``Web-scale k-means clustering'' \cite{sculley2010web}. 
Execution time for the Mini-Batch K-means implementation in scikit-learn on the same dataset was around 45 seconds for {\it K=600}. 
The running time of Mini-Batch K-Means against different values of $K$ is summarized in Table \ref{kmeans-time}
\begin{table}
\centering
\caption{Average execution time taken by MiniBatch K-Means for different values of $K$}
\label{kmeans-time}
\begin{tabular}{|l|l|}
\hline
K & Time(seconds) \\ \hline
100 &                7.898881 \\ \hline
200 &               12.972909 \\ \hline
300 &               17.126809 \\ \hline
400 &               20.611251 \\ \hline
500 &               23.639403 \\ \hline
600 &               45.750178 \\ \hline
700 &               33.864268 \\ \hline
800 &               55.053441 \\ \hline
900 &               74.374957 \\ \hline
1000 &               52.925832 \\ \hline
\end{tabular}
\end{table}

\begin{figure}
  \caption{Sample misclassified image 1}
  \label{fig:sample-misclass1}
  \centering
  \includegraphics[scale=0.5]{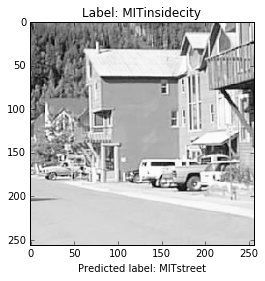}
\end{figure}

\begin{figure}
  \caption{Sample misclassified image 2}
  \label{fig:sample-misclass2}
  \centering
  \includegraphics[scale=0.5]{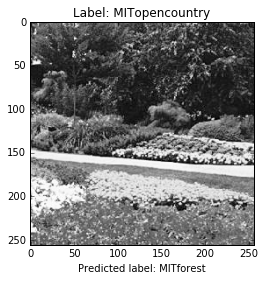}
\end{figure}

\begin{figure}
  \caption{Sample misclassified image 3}
  \label{fig:sample-misclass3}
  \centering
  \includegraphics[scale=0.5]{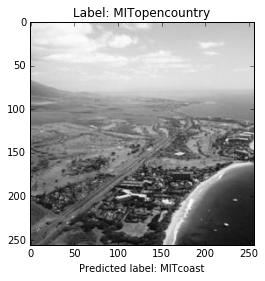}
\end{figure}

\begin{figure}
  \caption{Sample misclassified image 4}
  \label{fig:sample-misclass4}
  \centering
  \includegraphics[scale=0.5]{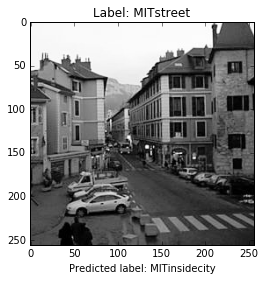}
\end{figure}

\begin{figure}
  \caption{Sample misclassified image 5}
  \label{fig:sample-misclass5}
  \centering
  \includegraphics[scale=0.5]{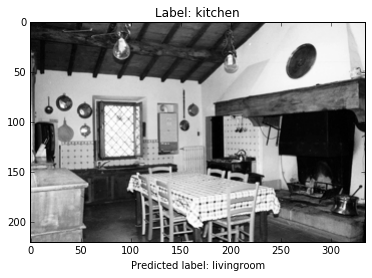}
\end{figure}

\begin{figure}
  \caption{Sample misclassified image 6}
  \label{fig:sample-misclass6}
  \centering
  \includegraphics[scale=0.5]{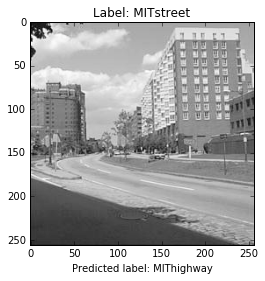}
\end{figure}

Some of the images misclassified by our model are shown in Figures 
\ref{fig:sample-misclass1}, \ref{fig:sample-misclass2}, \ref{fig:sample-misclass3}, \ref{fig:sample-misclass4}, 
\ref{fig:sample-misclass5} and \ref{fig:sample-misclass6}. It is interesting to note that some of these images may be 
prone to misclassification even for a human being.
\section{Conclusion}
\label{sec:conclusion}

We implemented an end-to-end scene recognition pipeline consisting of feature extraction, encoding,
pooling and classification steps using open-source libraries in python. We simultaneously utilized global feature descriptors as well as local feature descriptors 
from images, to form a hybrid feature descriptor for each image. DAISY descriptors associated with key points in images were 
utilized as local feature descriptor and HOG feature (at a different granularity) corresponding to an entire image was used as a global
feature descriptor. We made use of a bag of visual words encoding and applied Mini-Batch K-Means algorithm to 
significantly reduce the complexity of our encoding scheme. A 2-level pooling scheme was used to combine DAISY and HOG features 
corresponding to each image. Finally, a multi-class SVM classifier was trained with multiple kernel choices in a cross-validation 
setting to evaluate our model perfomance. The overall accuracy of our hybrid model was 76.4\% which was better
than the accuracies of models that made use of HOG or DAISY features alone. A full implementation of the proposed model 
is available in our github repository\cite{scenerecognition_github}.




%



\bibliographystyle{IEEEtran}
\bibliography{bare_conf}

\end{document}